%% file: anonymous-submission-latex-2024.tex
\documentclass[letterpaper]{article}
\usepackage[submission]{aaai24}
\usepackage{times}
\usepackage{helvet}
\usepackage{courier}
\usepackage[hyphens]{url}
\usepackage{graphicx}
\usepackage{natbib}
\usepackage{caption}
\usepackage{algorithm}
\usepackage{algorithmic}
\usepackage{newfloat}
\usepackage{listings}
\DeclareCaptionStyle{ruled}{labelfont=normalfont,labelsep=colon,strut=off}
\floatstyle{ruled}
\newfloat{listing}{tb}{lst}{}
\floatname{listing}{Listing}
\usepackage{booktabs}
\usepackage{multirow}

\title{ToolAlpaca: Generalized Tool Learning for Language Models \\ with 3000 Simulated Cases}

\author {
    Qiaoyu Tang\textsuperscript{\rm 1,3},
    Ziliang Deng\textsuperscript{\rm 1,3},
    Hongyu Lin\textsuperscript{\rm 1}\thanks{~ Corresponding Authors},
    Xianpei Han\textsuperscript{\rm 1,2}\footnotemark[1],\\
    Qiao Liang\textsuperscript{\rm 1,3},
    Boxi Cao\textsuperscript{\rm 1,3},
    Le Sun\textsuperscript{\rm 1,2}
}
\affiliations {
    \textsuperscript{\rm 1}Chinese Information Processing Laboratory ~
    \textsuperscript{\rm 2}State Key Laboratory of Computer Science \\
Institute of Software, Chinese Academy of Sciences, Beijing, China\\
\textsuperscript{\rm 3}University of Chinese Academy of Sciences, Beijing, China \\
{ \{tangqiaoyu2020,dengziliang2021,hongyu,xianpei\}@iscas.ac.cn} \\
{\{liangqiao2022,boxi2020,sunle\}@iscas.ac.cn}
}

\usepackage{bibentry}

\newcommand{\framework}{ToolAlpaca}
\newcommand{\dataLength}{3938}

\begin{document}

\maketitle

\begin{abstract}
Enabling large language models to utilize real-world tools effectively is crucial for achieving embodied intelligence. Existing approaches to tool learning have either primarily relied on extremely large language models, such as GPT-4, to attain generalized tool-use abilities in a zero-shot manner, or utilized supervised learning to train limited scopes of tools on compact models. However, it remains uncertain whether smaller language models can achieve generalized tool-use abilities without tool-specific training.
To address this question, this paper introduces ToolAlpaca, a novel framework designed to automatically generate a diverse tool-use corpus and learn generalized tool-use abilities on compact language models with minimal human intervention.
Specifically, ToolAlpaca first automatically creates a highly diversified tool-use corpus by building a multi-agent simulation environment. The corpus contains \dataLength{} tool-use instances from more than 400 real-world tool APIs spanning 50 distinct categories.
Subsequently, the constructed corpus is employed to fine-tune compact language models, resulting in two models, namely ToolAlpaca-7B and ToolAlpaca-13B, respectively.
Finally, we evaluate the ability of these models to utilize previously unseen tools without specific training.
Experimental results demonstrate that ToolAlpaca achieves effective generalized tool-use capabilities comparable to those of extremely large language models like GPT-3.5, demonstrating that learning generalized tool-use ability is feasible for compact language models. ~\footnote{Our code and data are available at \url{https://github.com/tangqiaoyu/ToolAlpaca}.}

\end{abstract}

\section{Introduction}

\input{sections/introduction}

\begin{figure*}[!tp]
  \centering
  \includegraphics[width=0.75\textwidth]{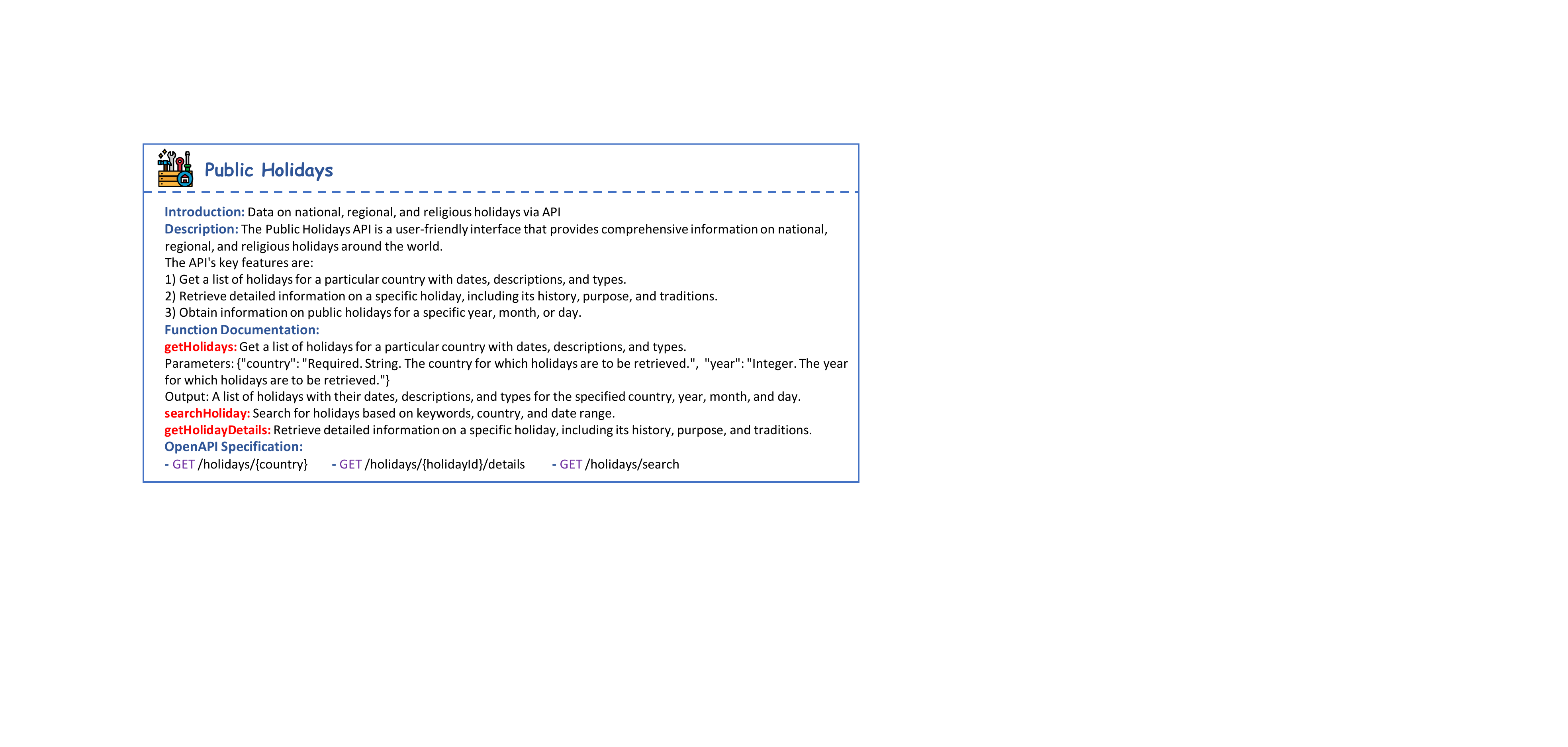}
  \caption{An instance of a tool documentation, composed of five essential parts: \textit{name, introduction, description, function documentation, OpenAPI specification}.}
  \label{fig:tool}
\end{figure*}
\section{Related Work}
\input{sections/related}

\section{Diversified Tool-use Corpus Generation via Multi-agent Simulation}

\input{sections/framework_v2}

\section{ToolAlpaca Corpus}
\input{sections/dataset}

\section{Experiment}
\input{sections/experiment}

\section{Conclusion}
\input{sections/conclusion}

\bibliography{aaai24}
\appendix
\input{sections/appendix}

\end{document}

%% file: sections/introduction.tex
Embodied intelligence, the ability to meaningfully interact with the environment, stands as a core attribute of advanced cognitive systems and a crucial advancement in artificial intelligence. The ability to create and use tools has expanded human beings' physical capabilities to interact with environments and augmented cognitive functions. Such evolutionary milestone has not only broadened our range of physical actions, but also brought about transformative changes in our problem-solving abilities and innovative thinking. The pursuit of incorporating tool-use capabilities into artificial intelligence holds great significance in advancing the development of general intelligent systems.

Recent advancements in enhancing large language models (LLMs) such as GPT-4 \citep{openai2023gpt4} with tool-use abilities have made significant progress in this area. These models have shown their ability to effectively employ external tools through integrated plugins, thereby expanding their versatility and enhancing the precision and quality of their outputs.
Unfortunately, due to a lack of understanding of how existing large language models acquire the general tool-use capability, currently compact language models still do not possess such general ability. Consequently, substantial research efforts are dedicated to fine-tuning smaller language models to acquire the capacity for tool usage~\citep{komeili-etal-2022-internet, DBLP:journals/corr/abs-2205-12255, DBLP:journals/corr/abs-2302-04761} on a limited range of tools, which lacks the ability to generalize to unseen tools. This discrepancy between the generalized tool-use abilities of larger models and the more constrained capabilities of compact models presents an intriguing question: \textit{Can these compact language models learn to generalize their tool-use abilities, thus enabling interaction with a broader spectrum of tools?}

In this paper, we explore whether it is feasible for compact language models to learn generalized tool-use abilities. Intuitively, previous studies have demonstrated the possibility of equipping compact language models with generalized instruction-following abilities by fine-tuning them on diversified instruction datasets~\citep{alpaca,zhou2023lima}.
Hence, a promising strategy for equipping language models with generalized tool-use abilities would involve fine-tuning them on a corpus containing highly-diversified tool-use instances. Unfortunately, such a diversified corpus is currently unavailable. This absence can be attributed to several crucial factors. First, the absence of a set of available tool APIs that can accommodate various tool usage scenarios for language models presents a considerable challenge in assembling a diverse collection of tools. Second, real-world tool-use instances often entail complex, intricate, and multi-turn interactions between the language model, users, and tools. This complexity significantly heightens the difficulty and manual effort involved in creating instances encompassing a wide array of tools on a large scale. Consequently, these factors significantly restrict the efforts to construct a diversified tool-use corpus for language model training efficiently.

\begin{figure}[t]
  \centering
  \includegraphics[width=0.49\textwidth]{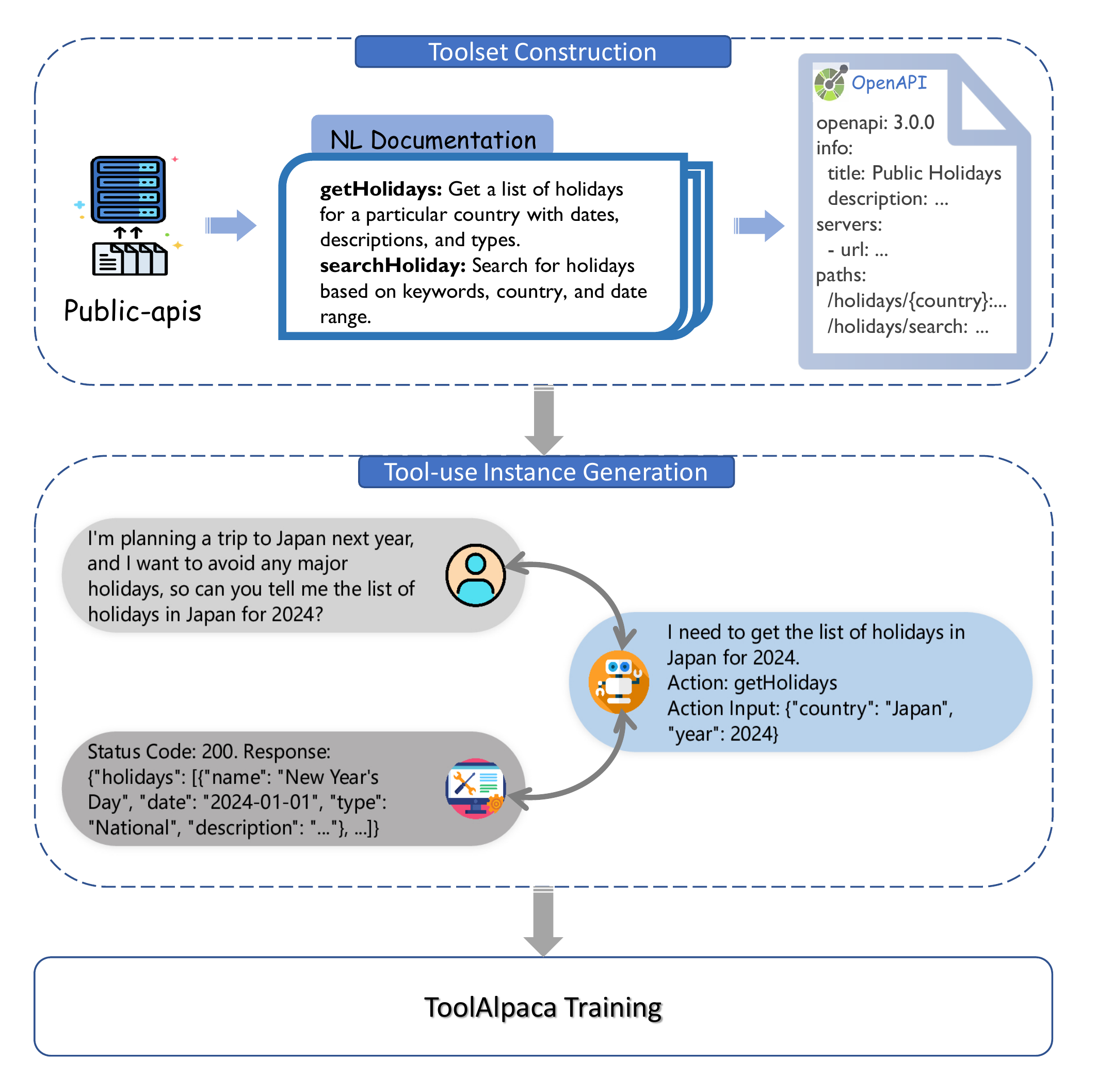}
  \caption{A high-level overview of ToolAlpaca, consisting of three components: (1)Toolset construction, where structured documentation for each tool is generated based on the brief introductions provided by public-apis. (2) Tool-use instance generation via multi-agent simulation. (3) ToolAlpaca model training, which involves fine-tuning language models on generated tool-use corpus to get ToolAlpaca.}
  \label{fig:pipeline}
\end{figure}

To this end, we propose a framework named \framework{}, which is designed to automatically create a diverse and well-structured toolset for LLMs and generate multi-turn complex tool-use instances for generalized tool learning. The overall structure of ToolAlpaca is shown in Figure~\ref{fig:pipeline}.
Specifically, to ensure the diversity and comprehensiveness of the toolset, \framework{} leverages LLM's text generation capability to construct a comprehensive toolset.
ToolAlpaca gathers a substantial amount of brief introductions of potentially valuable tools from the internet. It's important to note that there is no requirement for these tools' APIs to be functional or for them to possess structured documentation directly usable by LLMs.
Building on this foundation, ToolAlpaca employs the generative capacity of LLMs by taking the brief introduction of relevant tools as input and prompts the model to produce detailed, structured documentation for each tool.
By employing this methodology, ToolAlpaca has collected more than 400 tool descriptions spanning 50 categories. Each tool is uniformly represented using a standardized documentation format. Subsequently, in order to acquire tool-use instances involving the aforementioned tools, we have designed a simulation environment aimed at emulating the multi-step interactions among language models, users, and tools.
Specifically, we utilize LLMs to simulate the interactions between the model, users, and the APIs of the tools by leveraging LLMs to serve as different kinds of agents. In this way, our simulation environment can generate a substantial volume of tool-use instances without any manual intervention. Consequently, we have crafted an inclusive tool-use dataset that comprises \dataLength{} instances, effectively showcasing the practical application of over 400 distinct tools.

To verify whether our corpus can empower compact language models with the generalized tool-use ability, we conduct experiments to train ToolAlpaca model on Vicuna~\citep{vicuna2023}, a representative compact language model, and subsequently evaluate its performance on various unseen tools.
Through machine evaluation with GPT-4, we find that ToolAlpaca can effectively equip numerous unseen tools, ranging from real-world APIs to multi-modal tools, and it exhibits competitive performance with GPT-3.5.
Furthermore, we investigate the effect of diversity. It is observed that even with the same number of instances, the model trained on more varied toolsets will achieve better performance. This underscores that diversity is a pivotal factor for ToolAlpaca to generalize tool learning with 3000 simulated cases.

In summary, the main contributions of this paper are:
\begin{itemize}
    \item To the best of our knowledge, this paper is the first work that verifies the feasibility of equipping compact language models with generalized tool-use capacities as extremely large language models.
    \item This paper presents \framework{}, a simple framework for the automated generation of tool-use corpus and the enhancement of the compact language model's generalized tool-use ability.
    \item We create a diverse tool-use corpus containing 3.9k tool-use instances from more than 400 tools across 50 distinct categories. It serves as a solid foundation for compact language models to acquire generalized tool-use ability.
\end{itemize}

%% file: sections/related.tex
\paragraph{Tool Use} The utilization of external tools in LLMs has emerged as a rapidly growing research area~\citep{mialon2023augmented, qin2023tool}. 
Current approaches can be divided into two distinct categories.
The first category leverages the capabilities of LLMs, prompting them to interact with various tools, ranging from highly specialized ones such as code interpreters~\citep{2022arXiv221110435G, chen2022program}, search engines~\citep{yao2022react}, retrieval models~\citep{khattab2023demonstratesearchpredict} and AI models~\citep{shen2023hugginggpt, lu2023chameleon}, to more versatile toolsets~\citep{qin2023tool, li2023apibank, song2023restgpt}.
Large language models have already demonstrated robust generalization capabilities in tool usage and enable to equip numerous unseen tools via prompting.
In contrast, the second category concentrates on enhancing the tool-specific usage capabilities of compact language models through fine-tuning with datasets specifically designed for the specialized tools~\citep{DBLP:journals/corr/abs-2205-12255, DBLP:journals/corr/abs-2302-04761,xu2023tool}. 
Concurrent with our work, GPT4Tools~\citep{yang2023gpt4tools} fine-tuning compact models to incorporate multi-modal tools, which concentrates on a set of quite similar multi-modal tools. ToolLLM~\citep{qin2023toolllm} facilitates language models to master massive APIs. However, their data collection strategy requires the prior accumulation of massive authentic APIs, which requires manual efforts to obtain and verify.
Despite their effectiveness,  the domain of generalized tool-use abilities in compact language models remains largely unexplored upon the accomplishment of this paper.
This study aims to bridge this research gap by automatically constructing a diverse dataset on tool utilization that encompasses various tool-use scenarios.

\paragraph{LLMs for Data Generation}
Many research studies have employed LLMs for data generation, focusing on various tasks such as question answering~\citep{wang-etal-2021-want-reduce, agrawal2022qameleon, chen2023efficient},  semantic similarity predictions~\citep{schick-schutze-2021-generating}, and instruction tuning~\citep{honovich2022unnatural, wang2023selfinstruct}.
Furthermore, in the context of tool use, several works~\citep{DBLP:journals/corr/abs-2302-04761,patil2023gorilla,yang2023gpt4tools} have already employed model-synthesized data to enhance specific tool-use capabilities.
However, the generation of generalized tool-use data poses more significant challenges, as it involves extensive and diverse tools and more intricate multi-turn interactions.

%% file: sections/framework_v2.tex
In this section, we introduce \framework{}, a multi-agent simulation framework designed to generate a diversified tool-use corpus with minimal human intervention.
As shown in Figure~\ref{fig:pipeline}, our framework consists of two stages:
\begin{enumerate}
    \item Toolset Construction.
    This step aims to construct a collection of tools and represent them using a standardized format as \textit{\{name, introduction, description, function documentation, OpenAPI specification\}}.
    Specifically, we initiate the process by sourcing tool names and introductions from the internet and then utilize LLMs to enrich them with structured documentation that thoroughly delineates the functionality and usage of each tool.
    In this way, we can construct a diverse and structured toolset that closely resembles real-world scenarios.

    \item Tool-use Instance Generation.
    Given the toolset, this phase's objective is to generate tool-use instances within a simulation environment automatically.
    This environment is engineered through the orchestration of three distinct virtual agents, each embodied by a large language model: the user, the tool executor, and the assistant. 
    Through the multi-turn interplay among these agents, we can generate tool-use instances that reflect real-world tool utilization scenarios.
Each tool-use instance consists of three key elements: \textit{\{the user's instructions, the actions and their corresponding tool outputs, final response\}}.
\end{enumerate}

\subsection{Diverse Toolset Construction}

This section describes how to construct a diverse toolset and represent them in a uniform format.
The process initiates with the accumulation of an extensive API collection from the internet, reflecting real-world tool usage scenarios.
Given the rudimentary descriptions and lack of uniform representation in these APIs, we further leverage the generative capabilities of LLM to create comprehensive documentation for each tool. This documentation assists language models in understanding the functionality and usage of each tool.
Subsequently, we adhere to OpenAPI standards to generate a uniform specification for each API, enabling automated computer invocation and facilitating subsequent tool execution simulation.
In this way, each tool can be represented as a quintuple \textit{\{name, introduction, description, function documentation, OpenAPI specification\}}.
Figure~\ref{fig:tool} provides an example, where the name, description, and introduction offer basic information and the purpose of the public holiday tool, the function documentation provides the functionality, inputs and outputs of various functions (\textit{getHolidays, searchHolidays, getHolidayDetails}) contained within the tool, and the OpenAPI Specification provides a more comprehensive and structured document.
The detailed construction steps are elaborated as follows.

\paragraph{Tool Collection.} 
Various tools are commonly utilized by human beings, typically manifested in the form of web-based APIs. To facilitate the utilization and discovery of these APIs, a plethora of repositories exist on the Internet, aggregating a vast collection of practical and commonly used APIs.
Consequently, this step leverages the representative API repository, public-apis~\footnote{https://github.com/public-apis/public-apis}, as our target toolset. 
This repository encompasses over 1400 APIs spanning more than 50 distinct categories. From this, we collect the name and introduction of each tool.

\paragraph{Documentation Generation.}
To enhance the language model's comprehension of tools' functionalities and usage, this step employs LLMs to automatically generate documentation for each tool, including its description and function documentation.
Specifically, given the tool's name and introduction, we leverage the powerful generative capabilities of LLMs, such as ChatGPT, and construct corresponding prompts to generate the description and function documentation for each tool.
As illustrated in Figure~\ref{fig:tool}, for the tool description, we expand the introduction to provide a general overview of the tool's purpose and functionality, enabling the language model to understand the appropriate task scenarios for using the tool.
For the function documentation, we prompt LLMs to generate more specific functions within the scope described in the description, including their respective input and output parameters. It ensures that the LLM comprehends how to use the tool's different functionalities.
By employing this approach, we lay a solid foundation for subsequent user instruction generation and the creation of the simulated tool-use environment.

\paragraph{OpenAPI Specification Generation.}
Finally, we prompt LLM to generate a more formalized and structured representation for each tool in the form of OpenAPI Specification.
OpenAPI Specification (OAS) defines a standard and language-agnostic interface for describing APIs, including information about endpoints, expected input/output types, and possible error responses.
OAS provides consistency and readability for both humans and machines, making it an ideal choice for our tool-use corpus.
This comprehensive documentation serves as the foundation for simulating tool execution in the subsequent stages of our framework.

In this way, we construct a diverse, uniformly represented toolset, which provides a solid foundation for the multi-agent simulation environment building and further tool-use corpus generation.

\subsection{Automatic Tool-use Instances Generation}
Given the toolset, this section describes how to automatically construct a tool-use corpus, so that language models can be trained to acquire generalized tool-use ability.
Specifically, as depicted in Figure~\ref{fig:chat}, each tool-use instance can be represent as a triple \textit{\{Instruction, Actions, Response\}}:

\begin{figure}[!t]
  \centering
  \includegraphics[width=0.45\textwidth]{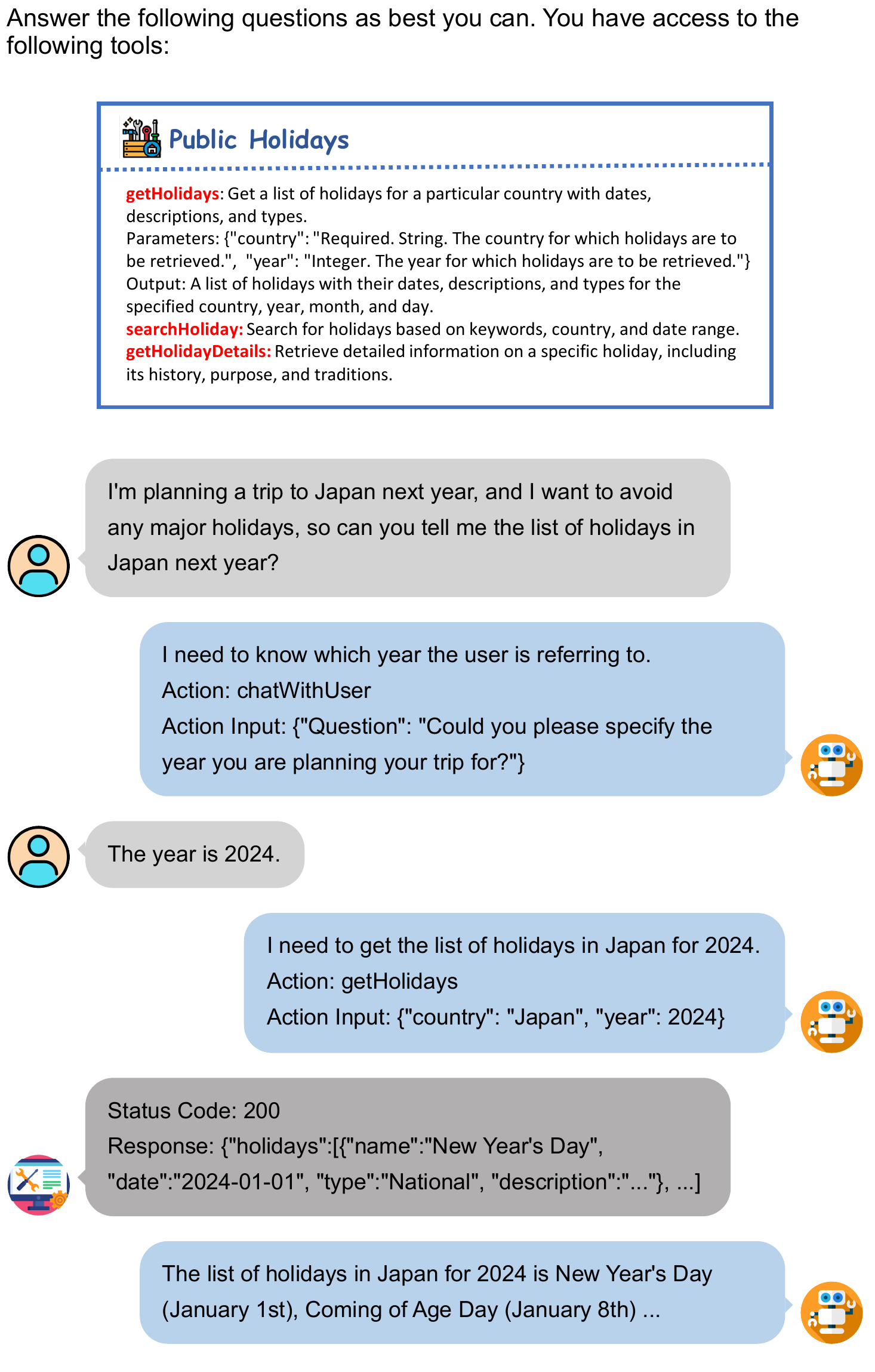}
  \caption{An illustration of the tool-use instance generation process within the simulation environment. The user agent initiates the sequence by providing an instruction. The assistant agent then interprets the instruction and engages in a multi-turn interaction with the user and the tool executor until a suitable response is generated.}
  \label{fig:chat}
\end{figure}

\begin{itemize}
    \item \textbf{Instruction:} A user query that requires tool assistance for resolution. \textit{"... so can you tell me the list of holidays in Japan next year?"} serves as an instruction in our example.

    \item \textbf{Actions:}
    The process of resolving an instruction may involve executing multiple actions in a specific order.
    Following React~\citep{yao2022react}, each action is represented by a tuple that includes the thought, the function name, the input parameters, and the corresponding tool response.
    For example, as shown in Figure~\ref{fig:chat}, the tuple \textit{("I need to get the list of holidays in Japan for 2024.", "getHolidays", \{"country": "Japan", "year": 2024\}, "Status Code: 200 Response:...")} represents an action.
    \item \textbf{Response:} This refers to the model's conclusive response after the multi-turn interaction, integrating the tool responses to provide a comprehensive solution to the original instruction. For instance, the response in our example is: \textit{"The list of holidays in Japan for 2024 is ..."}.
\end{itemize}

However, constructing a diverse and authentic tool-use dataset is a challenging task.
Firstly, the wide variety within our toolset makes it impracticable to manually draft instructions for each tool.
Given the vast array of tools, spanning from recreational to professional domains, and the fact that the construction of instructions relies on understanding the functionality and potential use cases of the tools, the burden of manual annotation becomes overwhelming.
Secondly, tool usage in real-world scenarios often involves a multi-round iterative process of trial and error, making the automated construction of tool-use instances that reflect real situations highly challenging.

To this end, we design a simulation environment to generate a tool-use corpus, encompassing three virtual agents: the user, the assistant, and the tool executor.
Tool-use instances are generated through the interplay among these agents.
Specifically, each agent is simulated by a large language model with a specific prompt. The distinct roles of each agent are detailed as follows:
\begin{itemize}

    \item \textbf{User Agent} is designed to mimic the tool user, with its functionalities encompassing: 
    (1) drafting task instructions for the current tool based on its function documentation;
    (2) responding to the assistant's queries based on the current interaction context, providing essential information that might be missing from the initial instruction.
    For each functionality, we construct corresponding prompt templates to guide LLMs to generate appropriate outputs.
    Moreover, to ensure diversity in task instructions, we have employed various prompts to generate instructions of different formats, including commands, questions, and others.
    Leveraging the large model's proficiency across virtually all domains, this approach enables the generation of high-quality and diversified instructions based on tool documentation. This effectively addresses the previously mentioned issues with manual annotation.

    \item \textbf{Assistant Agent} is designed to simulate an assistant with tool utilization capabilities. 
    It receives instructions from the user agent and determines the subsequent actions.
    This involves choosing the appropriate tools and functions, generating commands for the tool executor, and summarizing the interaction to generate the final response.
    As shown in Figure~\ref{fig:chat}, following ReAct~\citep{yao2022react}, we employ a (thought, action, observation) format template to guide LLM in accomplishing these tasks.

    \item \textbf{Tool Executor Agent} is constructed to emulate the execution of tools, receiving requests from the assistant agent and generating responses based on the tool's predefined functionalities.
    Specifically, after conducting format and parameter checks on the assistant's requests, these requests are converted into network request formats. Then the tool executor prompts LLM with the tool's OpenAPI specification and the assistant's requests to generate simulated execution results.
    Leveraging LLMs' robust simulation and generation capabilities, we mitigate the intricacies involved in constructing actual API calls. This method has been empirically validated as both accurate and effective, as evidenced in the following section.

\end{itemize}

Given the above agents, tool-use cases are generated through multiple rounds of interaction between them. 
Initially, the user agent generates instructions based on the tool information.
Subsequently, the assistant agent selects an appropriate action and its corresponding input and awaits simulation execution and response from the tool executor.
This iterative procedure of action selection and tool response collection continues until the assistant agent deems it has gathered sufficient information to respond to the user's instructions.
Through this multi-agent interaction, we can simulate realistic tool-use scenarios and generate comprehensive and diversified tool-use instances.

%% file: sections/dataset.tex
\subsection{Construction Details}
Leveraging the aforementioned multi-agent simulation framework, we have constructed the ToolAlpaca corpus. 
Specifically, the process begins with randomly selecting 500 APIs from the public-apis repository. Subsequently, we utilize ChatGPT to generate more comprehensive documentation, resulting in a varied and well-structured toolset.
Within our simulation environment, we use ChatGPT as the user agent to generate ten instructions for each tool, and the tool executor to simulate tool execution.
We appoint GPT-3.5 as the assistant agent due to its superior performance in structured output generation.\footnote{Preliminary experiments demonstrated an occasional inability of ChatGPT to strictly adhere to the prescribed output formats.}

To guarantee the quality of the data, we implement a simple yet effective filtering process on the generated corpus. We systematically exclude tools that might incorporate non-textual content within their inputs or outputs. In terms of instances, we discard those that exceed five interaction steps, lack relevant function calls, or exhibit parsing errors in their output.

Finally, we automatically construct an extensive and diversified tool-use corpus. As shown in Table \ref{tab:statistics}, it encompasses 426 distinctive tools from 50 categories, totaling \dataLength{} instances. In the following sections, we will analyze the diversity and quality of our corpus.

\begin{table}[!ht]
    \centering
    \resizebox{0.35\textwidth}{!}{
    \setlength{\tabcolsep}{12pt}
    \begin{tabular}{l c}
    \toprule
         statistics & ~ \\ \hline
         \# of Tool Categories & 50 \\
        \# of Tools & 426 \\
        \# of Instance & 3, 938 \\
         \quad  \# of single function call & 2, 512 \\
         \quad  \# of multiple function calls & 1, 426 \\
        avg. functions per tool & 4.85 \\
        avg. steps & 1.66 \\
        avg. instruction length & 23.42 \\
        avg. output length & 36.19 \\
    \bottomrule
    \end{tabular}}
    \caption{Statistics of ToolAlpaca corpus.}
    \label{tab:statistics}
\end{table}
\subsection{Diversity}
As previously underscored, diversity is pivotal for large models to acquire generalized capabilities and adapt to a multitude of unseen scenarios~\citep{wang2023selfinstruct}. ToolAlpaca corpus demonstrates diversity in two aspects:

\begin{figure}[!tp]
  \centering
  \includegraphics[width=0.4\textwidth]{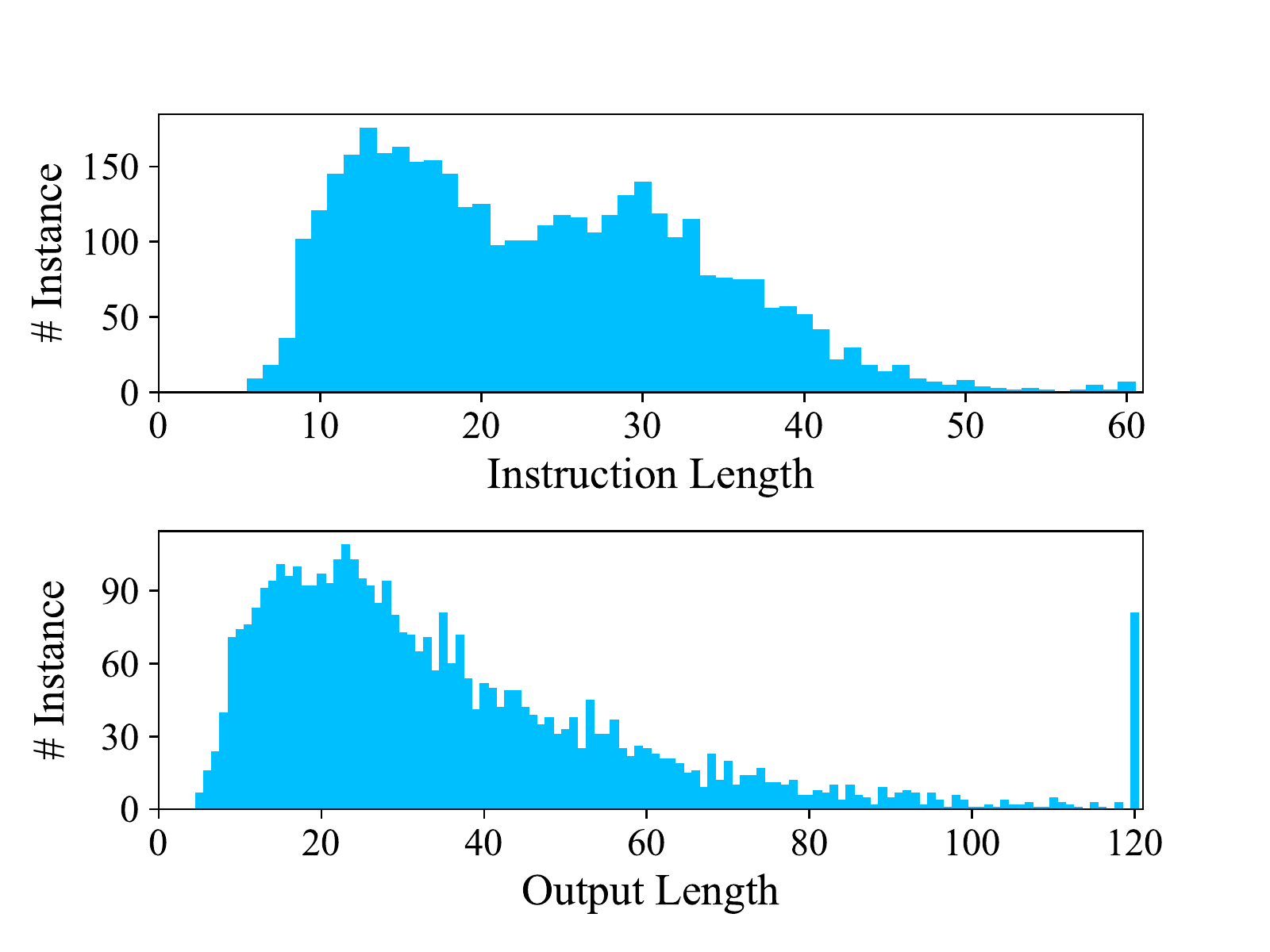}
  \caption{Length distribution of generated instructions and final outputs.}
  \label{fig:length}
\end{figure}
\begin{itemize}

    \item \textbf{Toolset.} As outlined in Table~\ref{tab:statistics}, our toolset demonstrates diversity in multiple aspects:
    (1) The toolset encompasses 50 categories of tools, ranging from common categories, such as jobs and news, to specialized categories like blockchain and finance.
    (2) Each tool provides an average of five functions, highlighting the diversity and comprehensiveness of its capabilities.
    (3) The range of function inputs varies from simple to complex scenarios, including arrays and objects, further enhancing the richness and complexity of our toolset.
    \item \textbf{Instances.} 
    The instances within the ToolAlpaca corpus demonstrate diversity in terms of instruction, function calls, and error handling.
    Specifically, we employ a variety of prompts during instruction generation to stimulate the language model in producing diverse instructions. The wide-ranging distribution of instruction length, as illustrated in Figure~\ref{fig:length}, partly substantiates this point.
    Additionally, our dataset contains about 1.5k instances that require multiple function invocations for resolution, further underscoring the comprehensiveness of our dataset.
    Furthermore, our data adequately reflects the potential errors that may be encountered in authentic tool usage scenarios, encompassing instances that involve various types of errors, such as invalid actions, parsing errors, and incorrect parameters.
\end{itemize}

\subsection{Quality}
\label{sec:quality}
To evaluate the quality of ToolAlpaca corpus, we randomly sample 100 instances and engage a human annotator for assessment.
The evaluation tests the solvability of the instructions generated by the user agent, the precision of the output from the tool executor agent, and the accuracy of the assistant agent's actions and responses.
As illustrated in Table~\ref{tab:quality}, we observe that the metrics for assessing the capabilities of the three agents all exceed 80\%. This substantiates that each agent is proficient in their respective roles, demonstrating the reliability of data constructed based on simulation and affirming the decent quality of our dataset.

\begin{table}[!ht]
    \centering
    \resizebox{0.43\textwidth}{!}{
    \begin{tabular}{l c}
    \toprule
         Quality & ~ Yes\%\\ \hline
        solvability of instructions & 88\% \\
        effectiveness of Tool agent's response & 92\% \\
        accuracy of action sequences and final output & 80\% \\
    \bottomrule
    \end{tabular}}
    \caption{Data quality review for ToolAlpaca corpus.}
    \label{tab:quality}
\end{table}

%% file: sections/experiment.tex
In this section, we investigate whether a set of simulated data can empower compact language models to acquire generalized tool-use capabilities.
To verify this, we conduct zero-shot experiments on various tools that have not appeared in the training set, ranging from simulated tools, real-world tools, to out-of-dataset multi-modal tools.
Furthermore, we investigate how the diversity of the toolset impacts the generalized tool-use ability of language models.

\subsection{Experimental Settings}
\paragraph{Training} We fine-tune Vicuna models (Vicuna-7B and Vicuna-13B) on ToolAlpaca corpus. The fine-tuning process consists of three epochs, with a batch size of 128 and a learning rate of 2e-5.

\begin{table*}[!ht]
    \centering
    \resizebox{0.75\textwidth}{!}{
    \begin{tabular}{l|c c c c|c c c}
    \toprule
        \multirow{2}{*}{Model} & \multicolumn{4}{c|}{Simulated Tools}  & \multicolumn{3}{c}{Real-world APIs} \\
        ~ & Procedure & Response & Overall & Human & Procedure & Response & Overall \\ \hline
        GPT-3.5 & 77.0 & 85.0 & 75.0 & 79.0 & 75.4 & 80.7 & 72.8 \\ \hline
        Vicuna-7B & 19.0 & 21.0 & 17.0 & 16.0 & 7.9 & 11.4 & 7.9 \\ 
        ToolAlpaca-7B & 63.0 & 69.0 & 60.0 & 73.0 & 63.2 & 57.9 & 55.3 \\ \hline
        Vicuna-13B & 17.0 & 31.0 & 16.0 & 25.0 & 13.2 & 16.7 & 12.3 \\ 
        ToolAlpaca-13B & 70.0 & 73.0 & 70.0 & 75.0 & 66.7 & 67.5 & 61.4 \\
    \bottomrule
    \end{tabular}}
    \caption{Evaluation results on unseen simulated tools and real-world APIs. We can observe that after training on our corpus, ToolAlpaca's performance significantly surpasses that of the Vicuna model, reaching comparable performance with GPT-3.5.}
    \label{tab:results}
\end{table*}

\paragraph{Evaluation}
To measure the generalized tool-use ability of the language model, we create an evaluation dataset through our data generation framework and manually annotate the data. This evaluation dataset consists of two subsets: (1) a simulated subset that includes 10 simulated tools, which were not part of the training toolset; (2) a real-world subset comprising 11 real-world APIs from various domains, designed to assess the divergence between our simulated data and real-world data.

To evaluate the models, we utilize GPT-4 for machine evaluation across all experiments, with an additional manual evaluation conducted specifically for the simulated subset. We prompt GPT-4 with the tool documentation and the standard answer from the human annotator and expect it to evaluate the performance in the following aspects:
\begin{itemize}
    \item \textbf{Procedure:} This metric evaluates the model's proficiency in accurately selecting suitable actions, utilizing correct parameters, and avoiding redundant actions.
    \item \textbf{Response:} This criterion measures whether the final response can satisfy the user's instruction. 
    \item \textbf{Overall:} This metric evaluates the whole process, requiring the correctness of procedure and response.
\end{itemize}

\subsection{Results}

\paragraph{Effectiveness of ToolAlpaca corpus.} 
Table \ref{tab:results} presents the main results from the simulated set, evidencing that fine-tuning on ToolAlpaca corpus can foster generalized tool learning for compact models. Without fine-tuning on our corpus, Vicuna models demonstrate constrained tool-use capabilities, with the human accept rate of 16 and 25, respectively. These statistics emphasize the existing compact models' insufficiency in achieving the generalized tool-use capacity like larger models. Nevertheless, our ToolAlpaca models can attain 73 (+57) and 75 (+50) accept rates, respectively. ToolAlpaca-13B even achieves comparable performance to GPT-3.5. This evidences the feasibility of instilling generalized tool-use capabilities into compact language models by only training on 3000 instances generated by our framework. Furthermore, the consistency between the human accept rate and the overall accuracy proves that machine evaluation can serve as a suitable evaluation method for this task.

\paragraph{Generalization on real-world tools.} 
 The effectiveness of our corpus is further validated through testing on real-world APIs, demonstrating that simulation serves as an exceptionally efficient data collection method.
 Table \ref{tab:results} exhibits the performance of ToolAlpaca on the real-world test set, where it achieves an overall accuracy of 55.3 and 61.4, respectively, significantly surpassing the performance of Vicuna models. This suggests that training on simulated data can indeed adapt to real-world tool usage scenarios. We attribute this to the current LLMs' robust simulation capabilities, which provide compelling evidence for future simulation-based data construction.

\begin{table}[!ht]
    \centering
    \resizebox{0.4\textwidth}{!}{
    \begin{tabular}{l c c c c}
    \toprule
        Model & $SR_t$ & $SR_{act}$ & $SR_{args}$ & $SR$ \\ \hline
        GPT-3.5 & 99.5 & 99.5 & 91.5 & 91.5 \\ \hline
        Vicuna-13B & 84.4 & 43.7 & 46.7 & 26.2  \\ 
        GPT4Tools & 98.2 & 97.0 & 92.2 & 90.6 \\
        ToolAlpaca-13B* & - & 95.5 & 85.3 & 83.7 \\
    \bottomrule
    \end{tabular}}
    \caption{Evaluation results on unseen tools from GPT4Tools test set. Metrics: successful rate of thought, action, arguments, and the entire instance. We can observe that ToolAlpaca, with 3.9k cases, reaches the same performance level as GPT4Tools, which has been trained on 71k instances generated by the same process with the test set. *: As our training set does not include data not involving tool use, we exclude 50 out of 652 test cases that do not involve tool usage.}
    \label{tab:gpt4tools}
\end{table}

Moreover, to evaluate ToolAlpaca's generalization on out-of-dataset scenarios, we conduct experiments on GPT4Tools\citep{yang2023gpt4tools} test set, which encompasses 8 multi-modal tools.
As shown in Table~\ref{tab:gpt4tools},  ToolAlpaca, trained on merely 3.9k cases, demonstrate 83.7 success rate on out-of-dataset evaluation, which is close to GPT4Tools, trained on 71k instances constructed with the same process.
 This observation indicates that the language model can invoke out-of-dataset tools after training on ToolAlpaca corpus. 
 We speculate that the performance may be attributed to the diversity of instances and toolset, and we delve into it in the subsequent experiment.

\begin{figure}[!tp]
  \centering
  \includegraphics[width=0.3\textwidth]{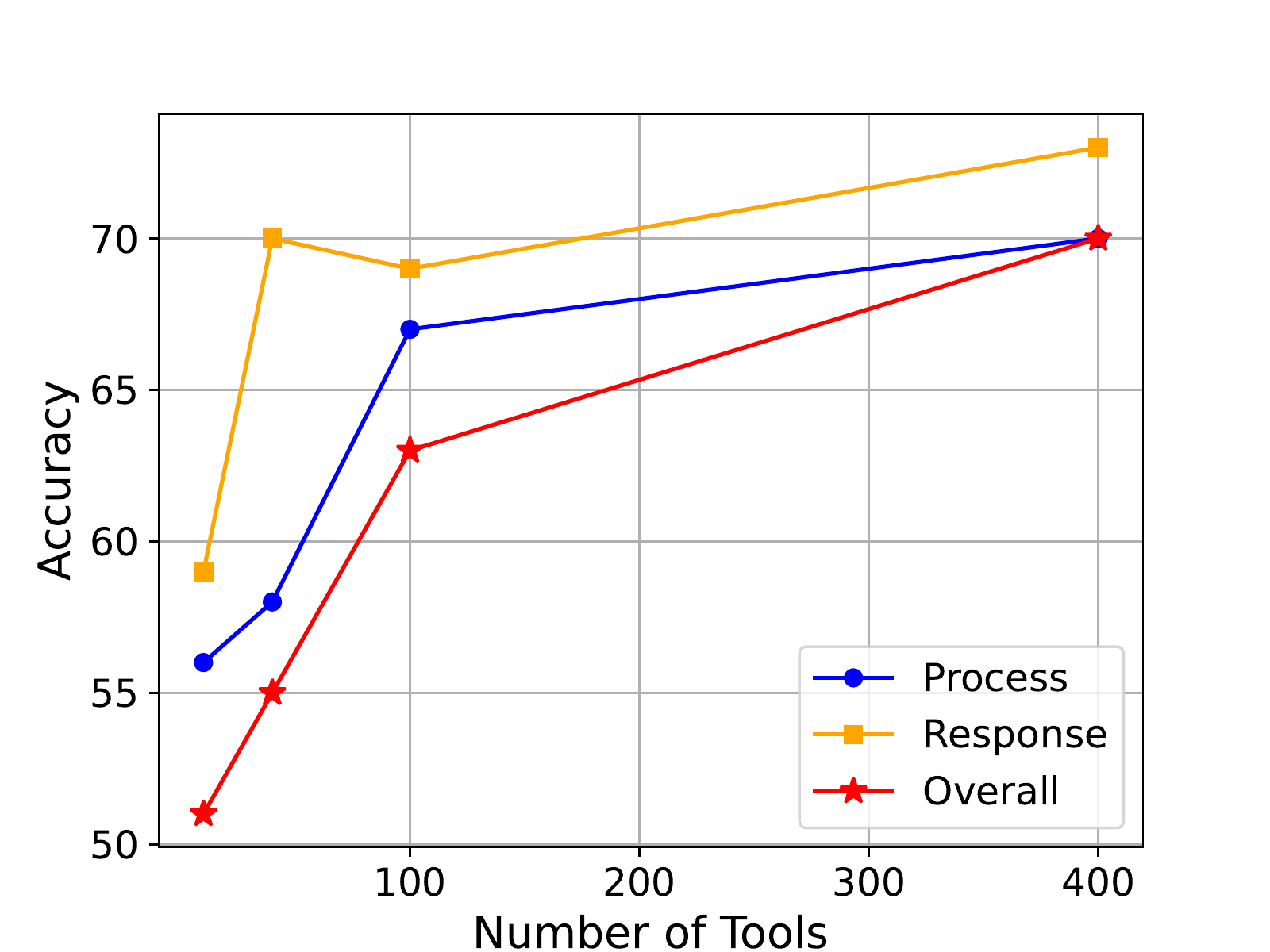}
  \caption{Performance variation with the increasing toolset diversity.}
  \label{fig:ablation}
\end{figure}

\paragraph{Impact of diversity.} 
 The diversity of the dataset is crucial for the generalization of tool learning.
 To investigate this, we maintain the number of instances and construct datasets on 10, 40, 100, and 400 tools, respectively.
 Subsequently, we fine-tune Vicuna-13B on these datasets with the same experimental settings and utilize GPT-4 to evaluate the validation set.
 As shown in Figure~\ref{fig:ablation}, as the diversity of the toolset increases, the performance on the validation set gradually improves.
Specifically, training with a dataset of 10 different tools resulted in a mere 51 overall accuarcy. In contrast, when the variety of tools increases to 400 and keeps the number of instances, the performance escalates to 70. 
This finding highlights the significant role of toolset diversity in generalizing tool learning.
This provides valuable insight for the construction of datasets for generalized ability learning.

%% file: sections/conclusion.tex
In this paper, we introduce \framework{}, an automated framework designed to improve the generalized tool-use capability of language models.
Specifically, we first create a comprehensive corpus spanning a broad range of tools with various usage instances. 
Subsequently, this corpus serves as the basis for fine-tuning compact language models, leading to the generation of the ToolAlpaca models. 
Experimental results indicate that ToolAlpaca performs comparably to GPT-3.5 in generalized tool-use scenarios. This finding not only substantiates the potential of our data generation framework but also highlights the feasibility of mastering generalized tool use in compact-size models.

%% file: sections/appendix.tex
\newpage

\section{Implementation Details}
\begin{figure*}[!t]
  \centering
  \includegraphics[width=0.8\textwidth]{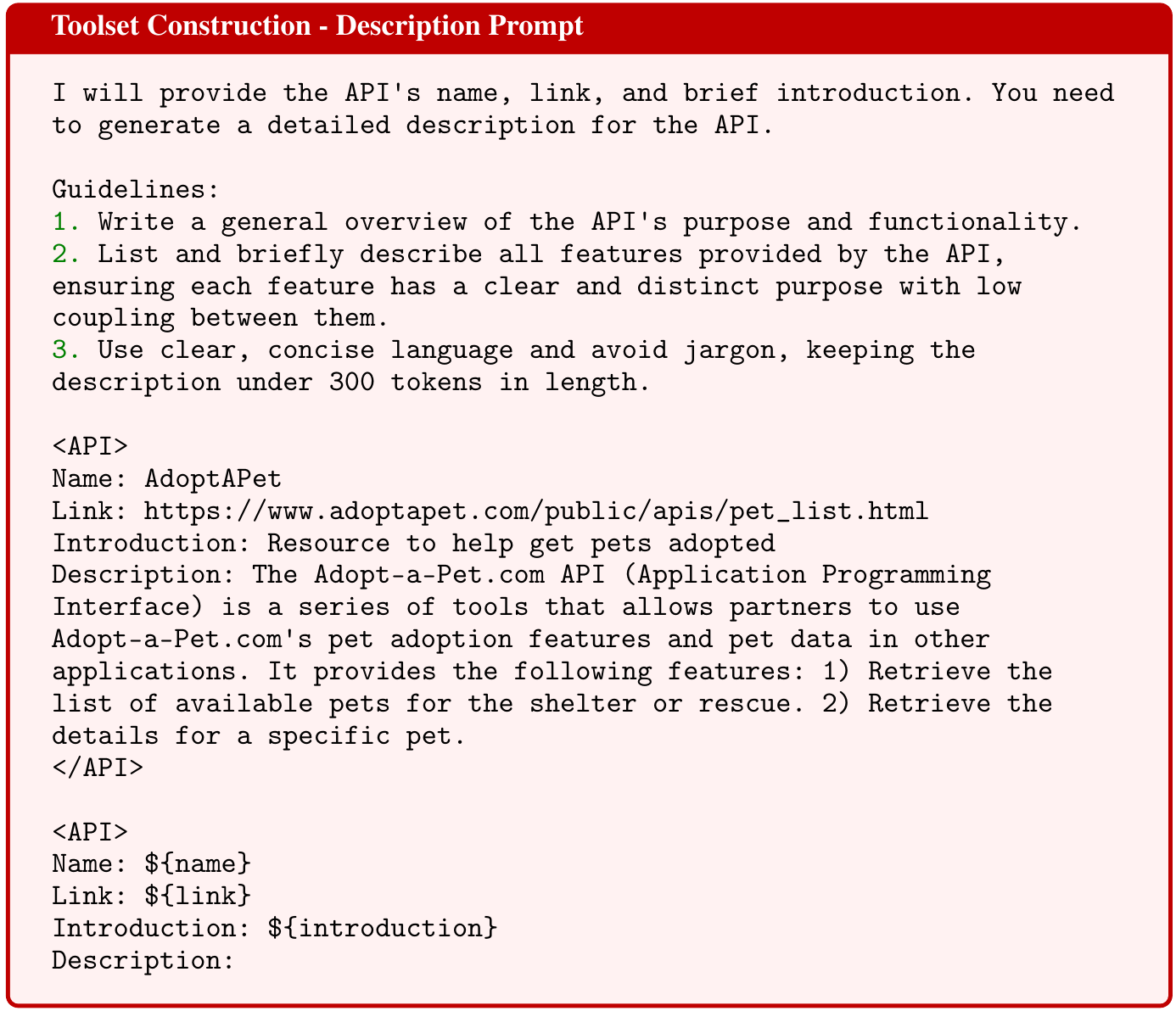}
  \caption{Description generation prompt.}
  \label{fig:description}
\end{figure*}

\begin{figure*}[!t]
  \centering
  \includegraphics[width=0.8\textwidth]{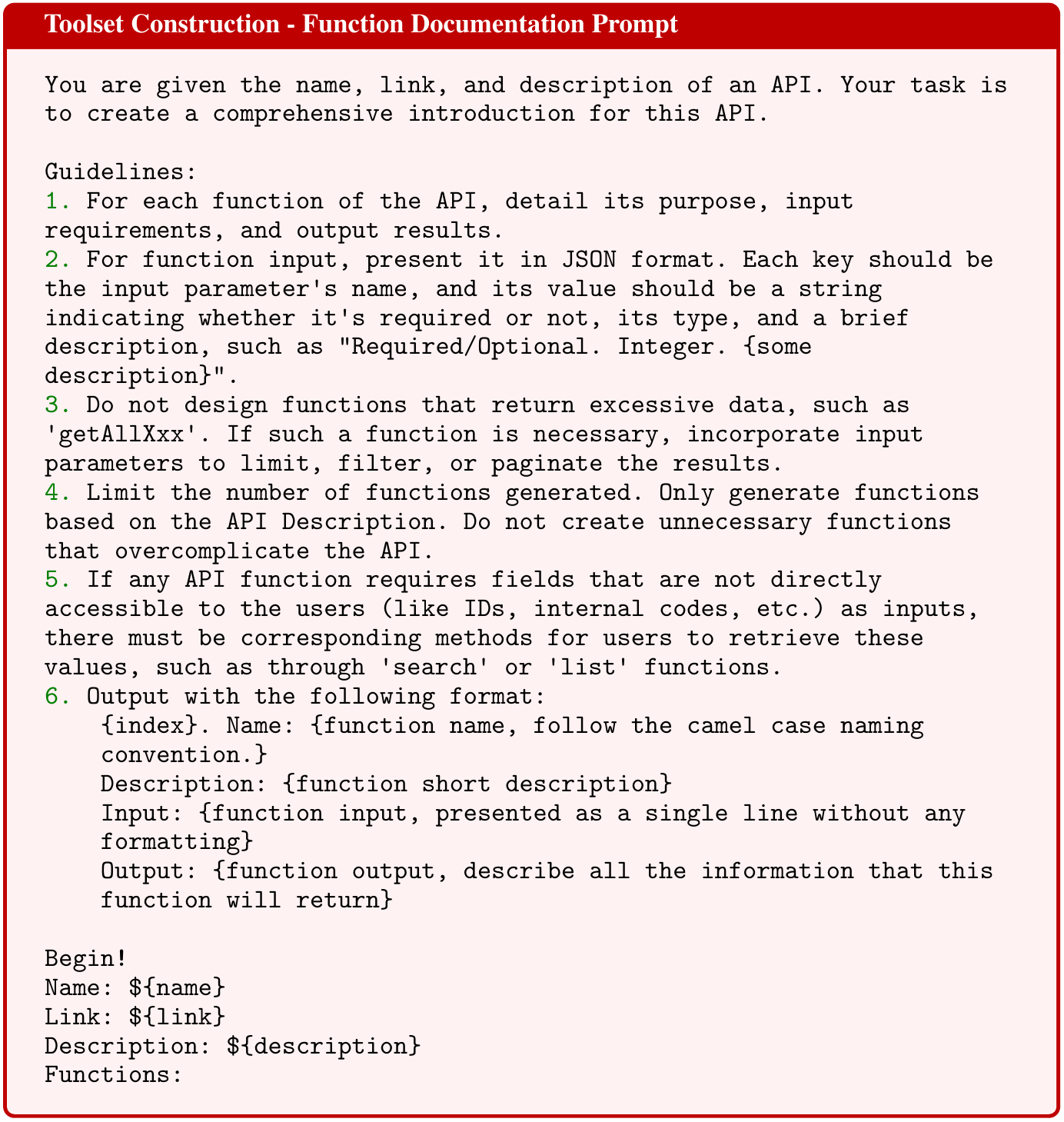}
  \caption{Function documentation generation prompt.}
  \label{fig:function}
\end{figure*}

\begin{figure*}[!t]
  \centering
  \includegraphics[width=0.8\textwidth]{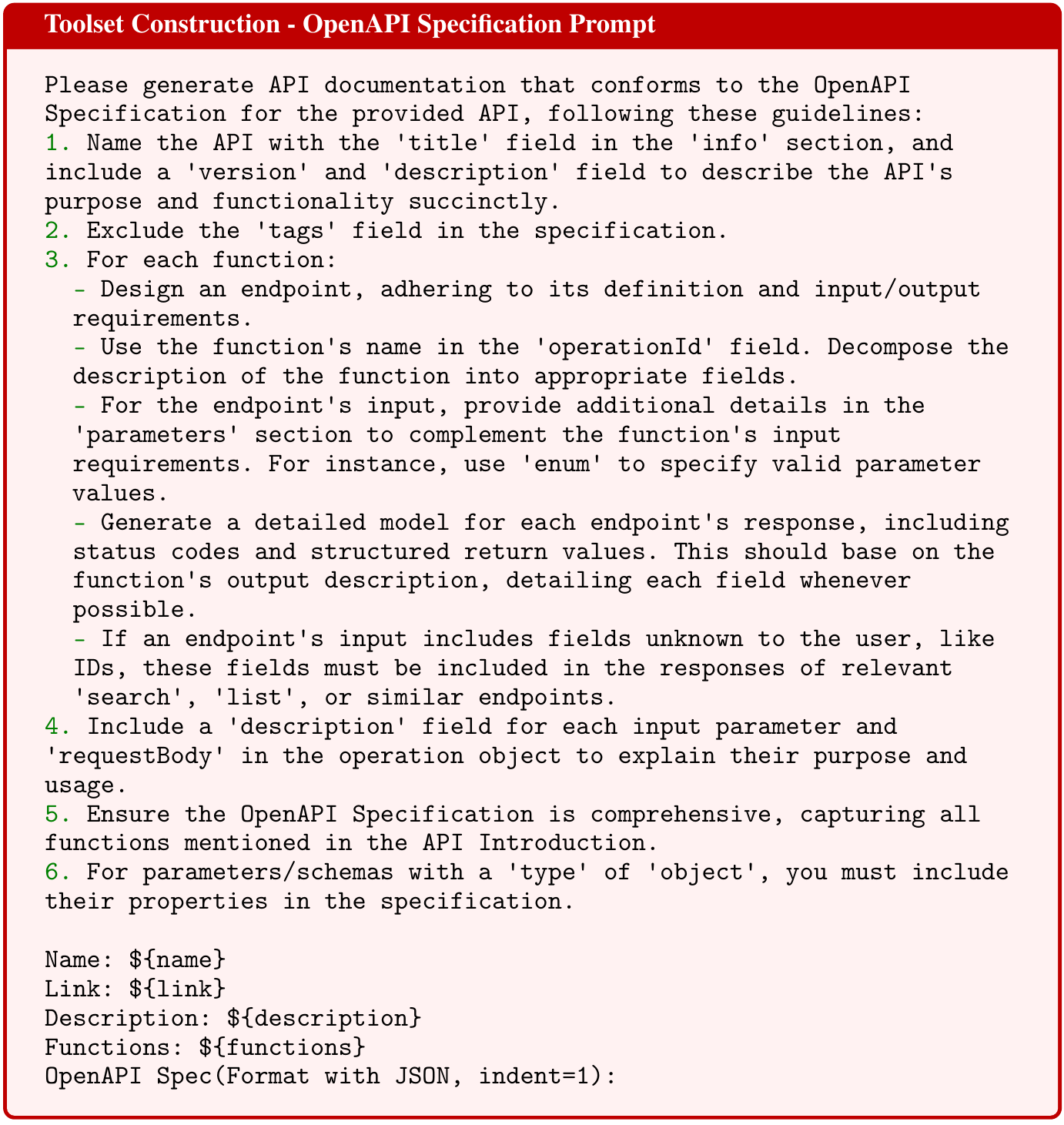}
  \caption{Openapi specification generation prompt.}
  \label{fig:openapi}
\end{figure*}

\begin{figure*}[!t]
  \centering
  \includegraphics[width=0.8\textwidth]{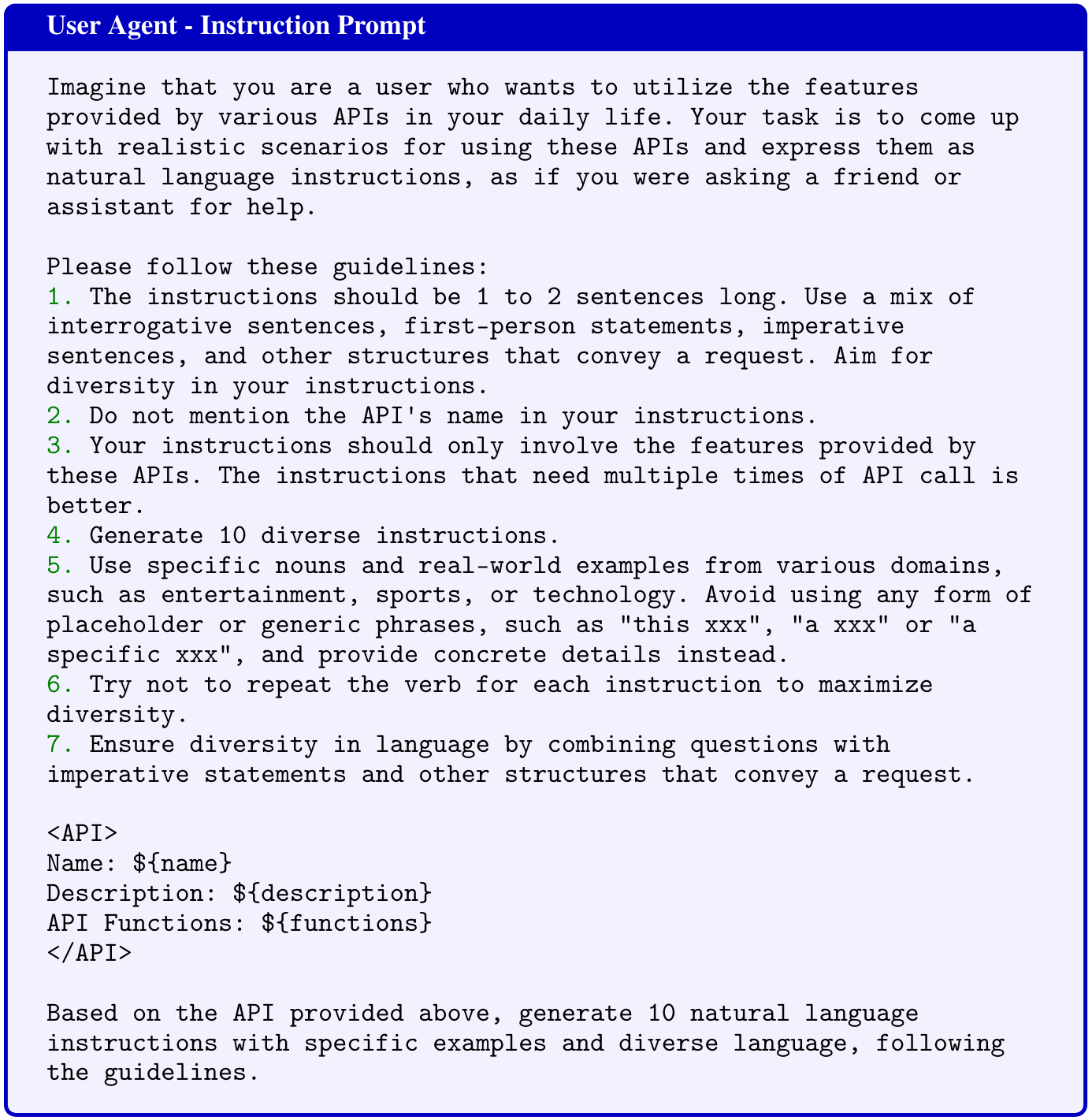}
  \caption{User agent prompt 1 for instruction generation.}
  \label{fig:user1}
\end{figure*}

\begin{figure*}[!t]
  \centering
  \includegraphics[width=0.8\textwidth]{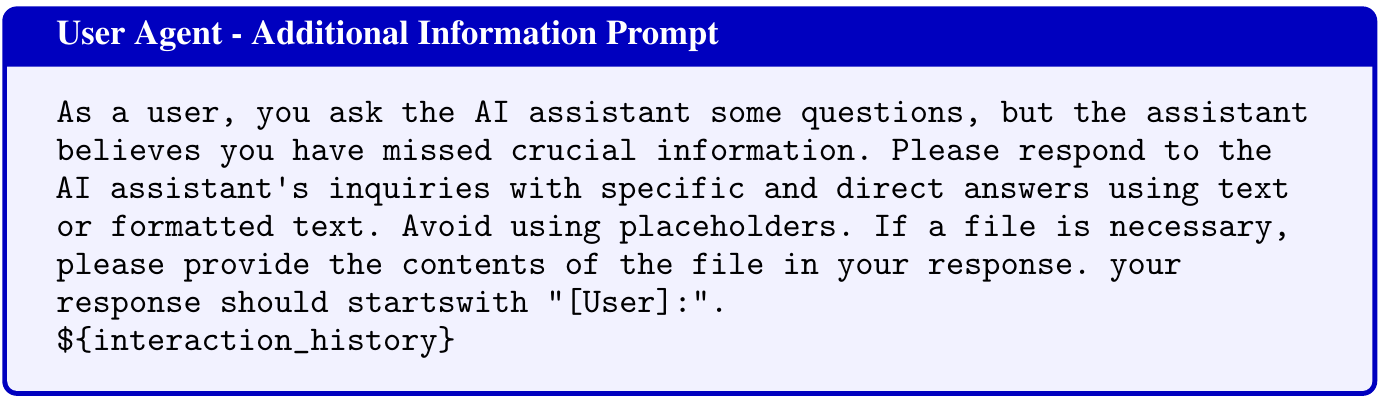}
  \caption{User agent prompt 2 for providing missing information.}
  \label{fig:user2}
\end{figure*}

\begin{figure*}[!t]
  \centering
  \includegraphics[width=0.8\textwidth]{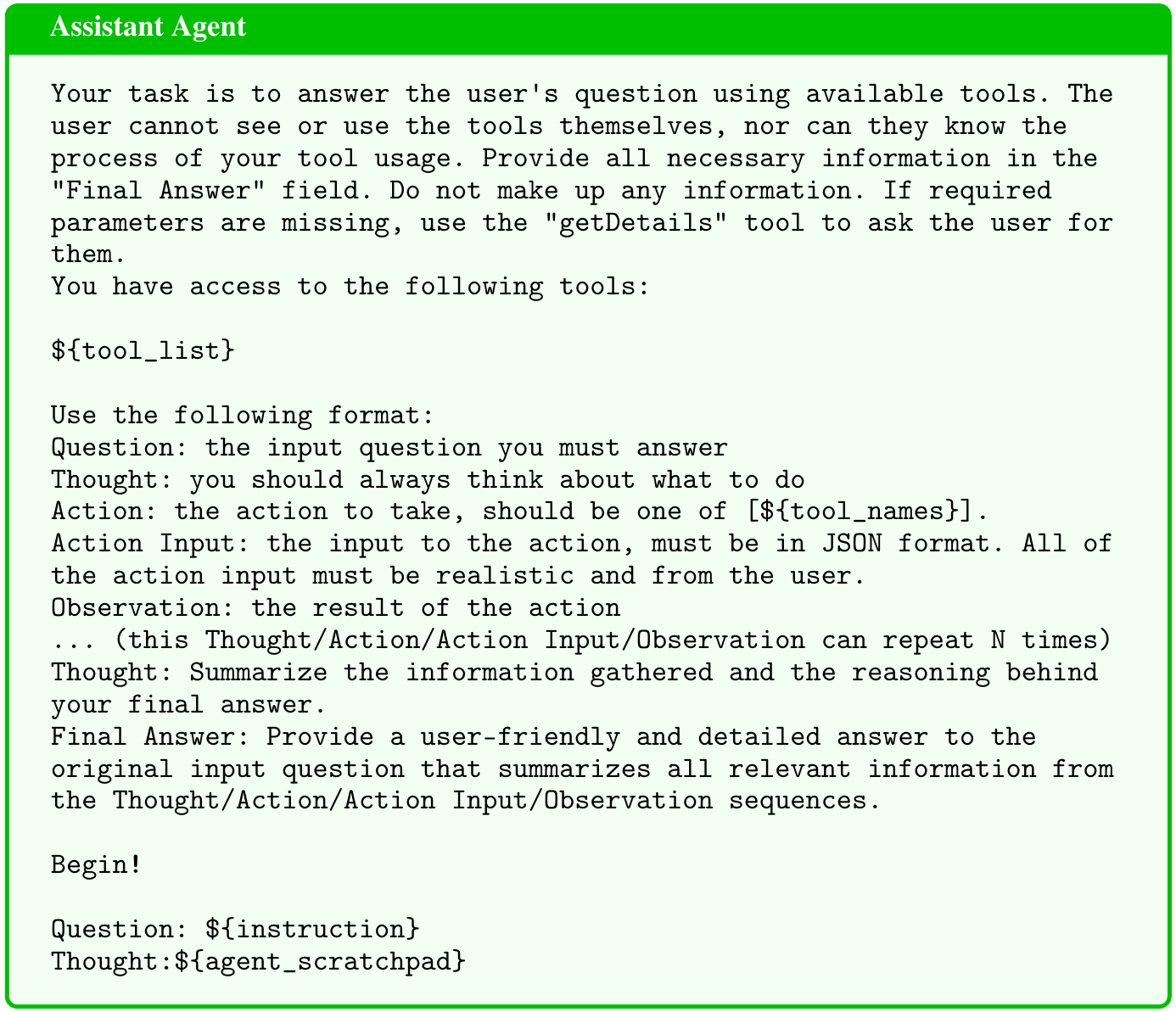}
  \caption{Assistant agent prompt.}
  \label{fig:assistant}
\end{figure*}

\begin{figure*}[!t]
  \centering
  \includegraphics[width=0.8\textwidth]{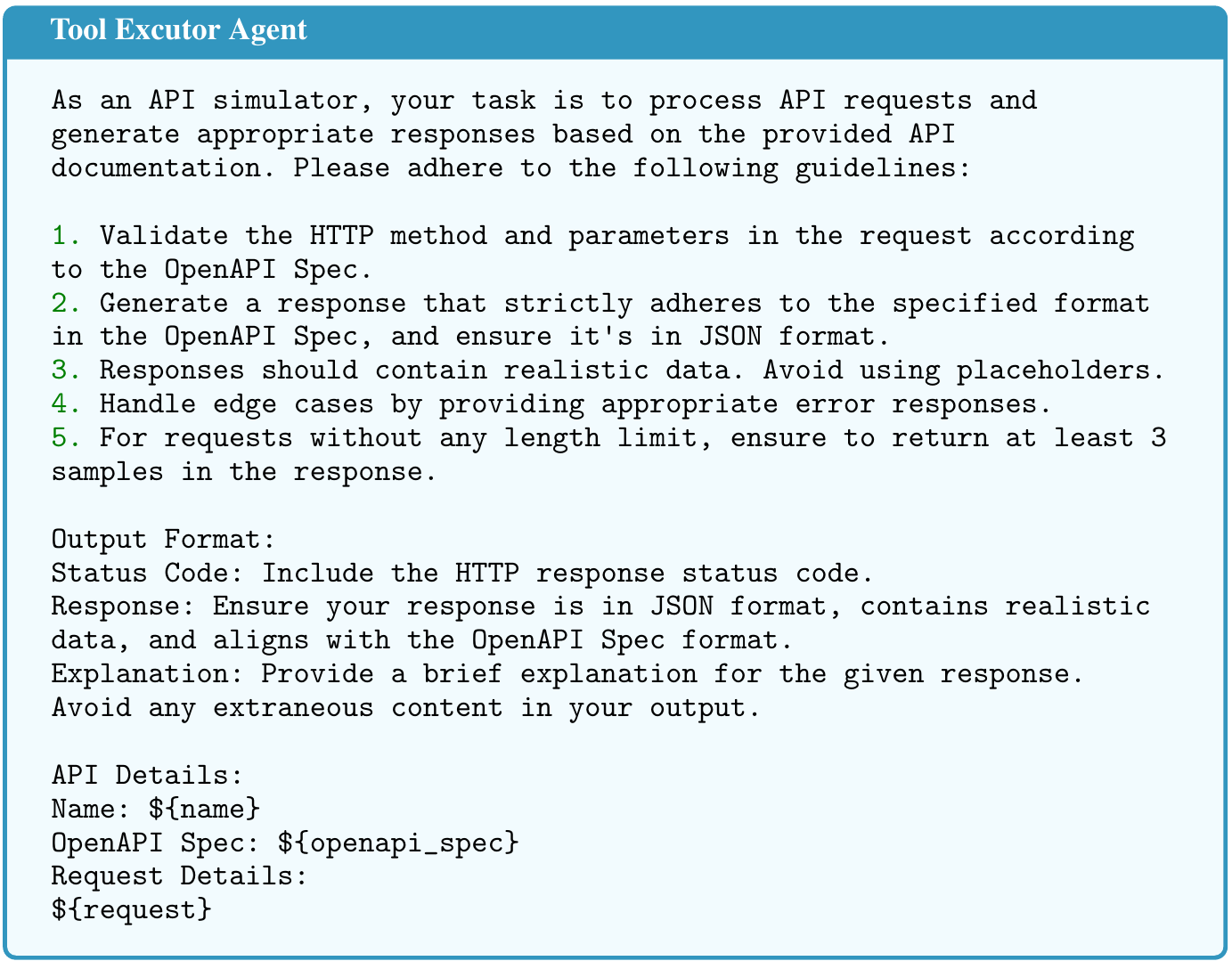}
  \caption{Tool executor agent prompt.}
  \label{fig:tool_executor}
\end{figure*}

In this section, we show the details of prompt templates in ToolAlpaca. Figure~\ref{fig:description}, Figure~\ref{fig:function}, and Figure~\ref{fig:openapi} delineate the prompts employed during toolset construction. Figure~\ref{fig:user1} and Figure~\ref{fig:user2} illustrate the corresponding prompts for user agent's two responsibility, generating user instructions and providing missing information.
The prompts designed for to the assistant agent and the tool executor agent are detailed in
Figure~\ref{fig:assistant} and Figure~\ref{fig:tool_executor}. \textit{"\$\{...\}"} within the prompts are placeholders, will be replaced by real variables during the generation process.

\section{Experiment Details}
\subsection{Hyperparameters}
The fine-tuning configuration for ToolAlpaca is recorded in Table~\ref{tab:params}.
\begin{table}[!ht]
    \centering
    \resizebox{0.3\textwidth}{!}{
    \begin{tabular}{l c}
    \toprule
        Hyperparameters & Value \\ \hline
        optimizer &  AdamW \\ 
        learning rate &  2e-5 \\
        weight decay & 0.0 \\
        warmup ratio &  0.03\\
        lr scheduler type & cosine \\
        num train epochs &  3 \\
        batch size & 128 \\
        max length & 2048 \\
    \bottomrule
    \end{tabular}}
    \caption{The fine-tuning configuration for ToolAlpaca.}
    \label{tab:params}
\end{table}
\subsection{Evaluation Dataset Details}
To evaluate the generalized tool-use ability, we construct the evaluation dataset via our framework, which consists two subsets: a simulated subset with 10 simulated tools and 100 instances, a real-world subset with 11 real-world APIs and 114 instances. The toolset used in the evaluation datset is detailed in Table~\ref{tab:tool}.

\begin{table*}[!ht]
    \centering
    \resizebox{0.98\textwidth}{!}{
    \begin{tabular}{l c l}
    \toprule
        Name & Category & Introduction \\ \hline
        \textit{Simulated Tools} & ~ & ~\\ \hline
       Axolotl & Animals & Collection of axolotl pictures and facts  \\ 
        AniAPI & Anime & Anime discovery, streaming \& syncing with trackers  \\ 
        AbuseIPDB & Anti-Malware & IP/domain/URL reputation  \\ 
        Améthyste & Art \& Design & Generate images for Discord users  \\ 
        Auth0 & Authentication \& Authorization & Easy to implement, adaptable authentication and authorization platform  \\ 
        Abstract Public Holidays & Calendar & Data on national, regional, and religious holidays via API  \\ 
        1Forge & Currency Exchange & Forex currency market data  \\ 
        A Bíblia Digital & Books & Do not worry about managing the multiple versions of the Bible  \\ 
        Apache Superset & Business & API to manage your BI dashboards and data sources on Superset  \\ 
        Lob.com & Data Validation & US Address Verification \\\hline
        \textit{Real-world APIs} & ~ & ~\\ \hline
        Nager.Date & Calendar & Public holidays for more than 90 countries  \\ 
        airportsapi & Transportation & Get name and website-URL for airports by ICAO code  \\ 
        AviationAPI & Transportation & FAA Aeronautical Charts and Publications, Airport Information, and Airport Weather  \\ 
        chucknorris.io & Entertainment & JSON API for hand curated Chuck Norris jokes  \\ 
        Random Useless Facts & Entertainment & Get useless, but true facts  \\ 
        apilayer weatherstack & Weather & Real-Time \& Historical World Weather Data API  \\ 
        Free Dictionary & Dictionaries & Definitions, phonetics, pronounciations, parts of speech, examples, synonyms  \\ 
        WolframAlpha & Machine Learning & Provides specific answers to questions using data and algorithms  \\ 
        Fruityvice & Food \& Drink & Data about all kinds of fruit  \\ 
        Cataas & Animals & Cat as a service (cats pictures and gifs)  \\ 
        CurrencyBeacon & Currency Exchange & Real-time and historical currency rates JSON API \\
    \bottomrule
    \end{tabular}}
    \caption{Tools used in our evaluation dataset.}
    \label{tab:tool}
\end{table*}

\subsection{Evaluation Prompt}
\begin{figure*}[!t]
  \centering
  \includegraphics[width=0.8\textwidth]{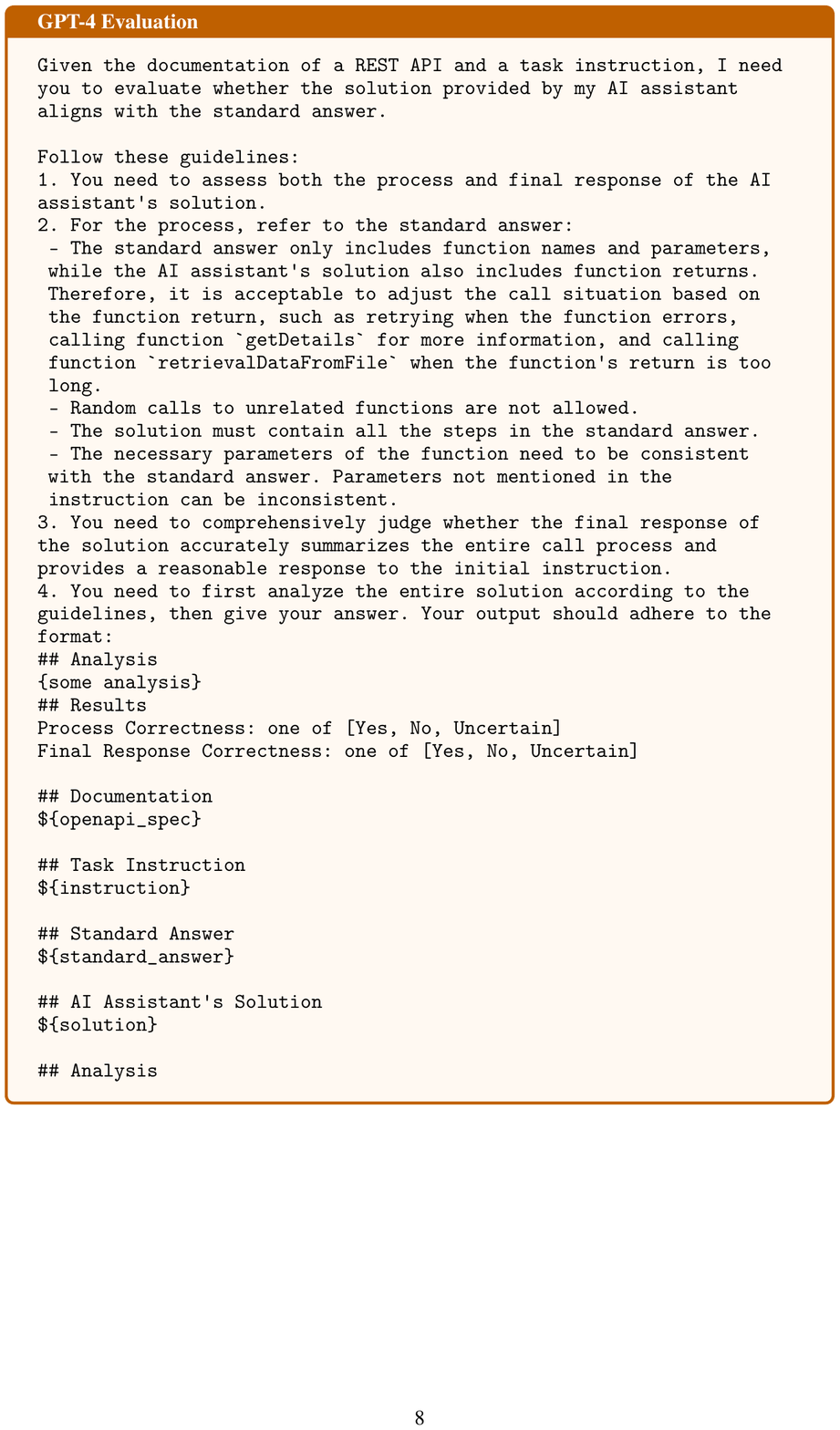}
  \caption{Evaluation prompt for GPT-4.}
  \label{fig:eval}
\end{figure*}

Following the evaluation method used by Vicuna~\citep{vicuna2023}, we use GPT-4 as our evaluator. The evaluation prompt is shown in Figure~\ref{fig:eval}.
\subsection{Case Study}
\begin{figure*}[!t]
  \centering
  \includegraphics[width=0.8\textwidth]{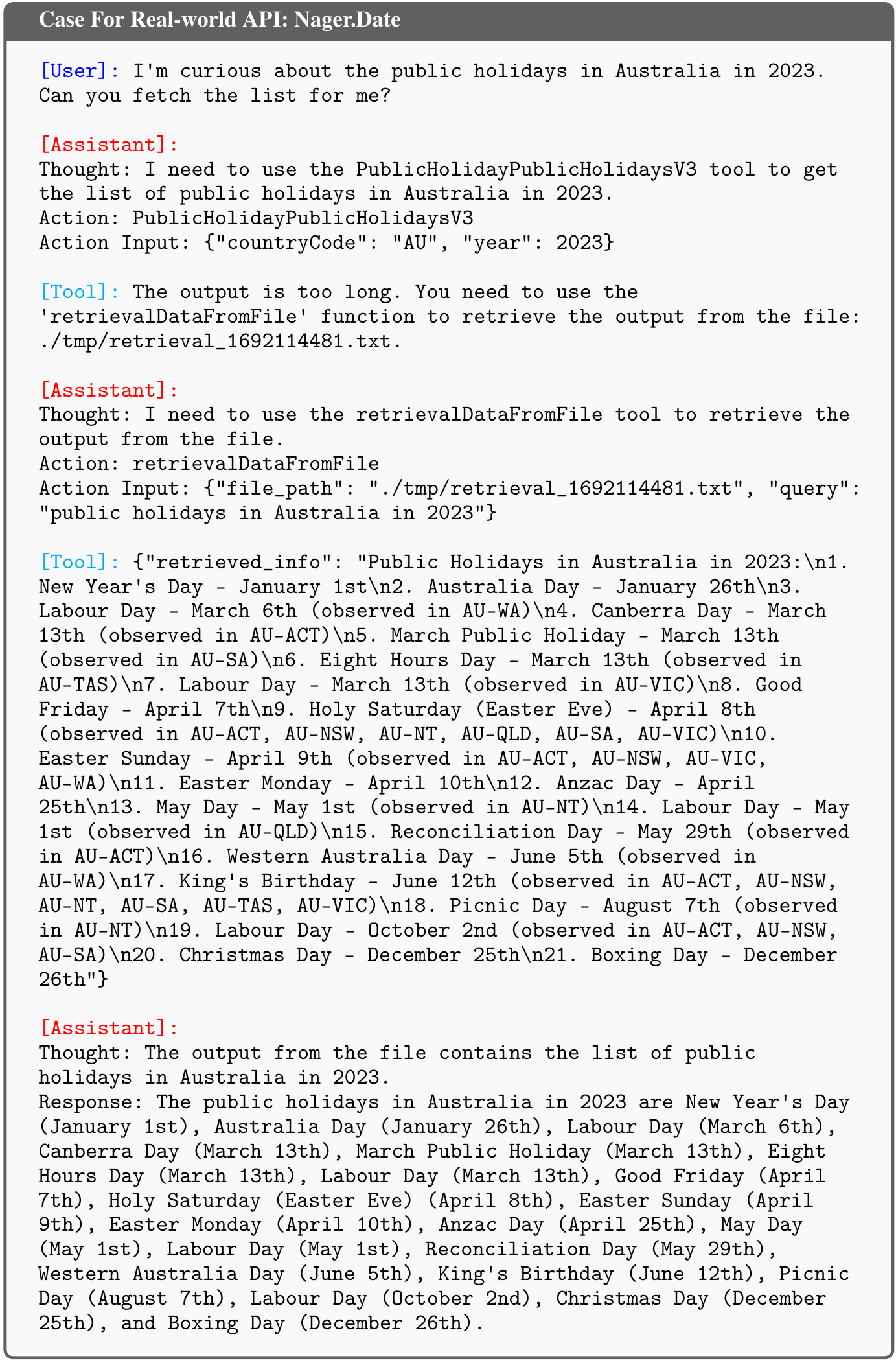}
  \caption{Cases for real-world API: Nager.Date.}
  \label{fig:case1}
\end{figure*}

\begin{figure*}[!t]
  \centering
  \includegraphics[width=0.8\textwidth]{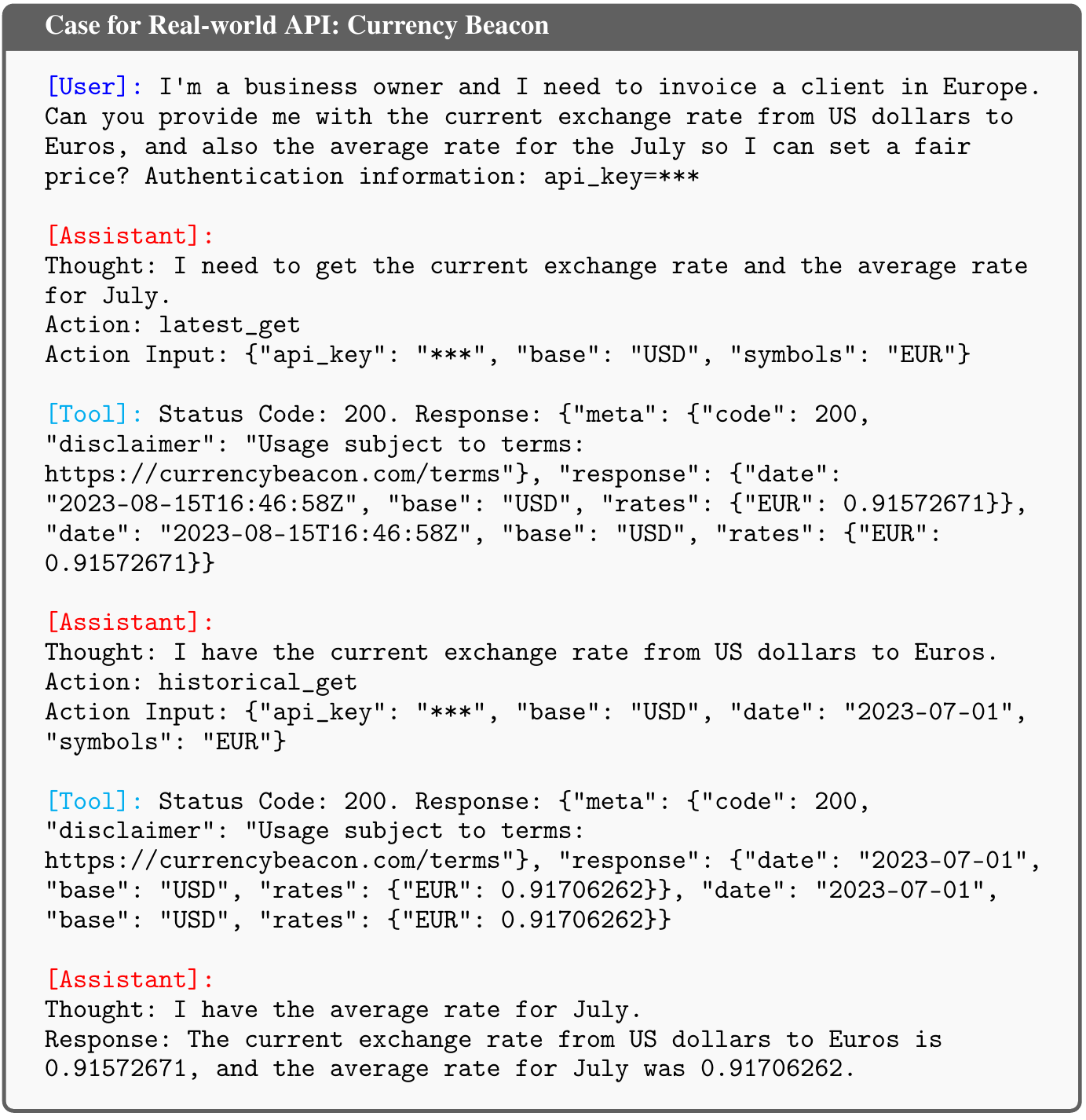}
  \caption{Cases for real-world API: Currency Beacon.}
  \label{fig:case2}
\end{figure*}

\begin{figure*}[!t]
  \centering
  \includegraphics[width=0.8\textwidth]{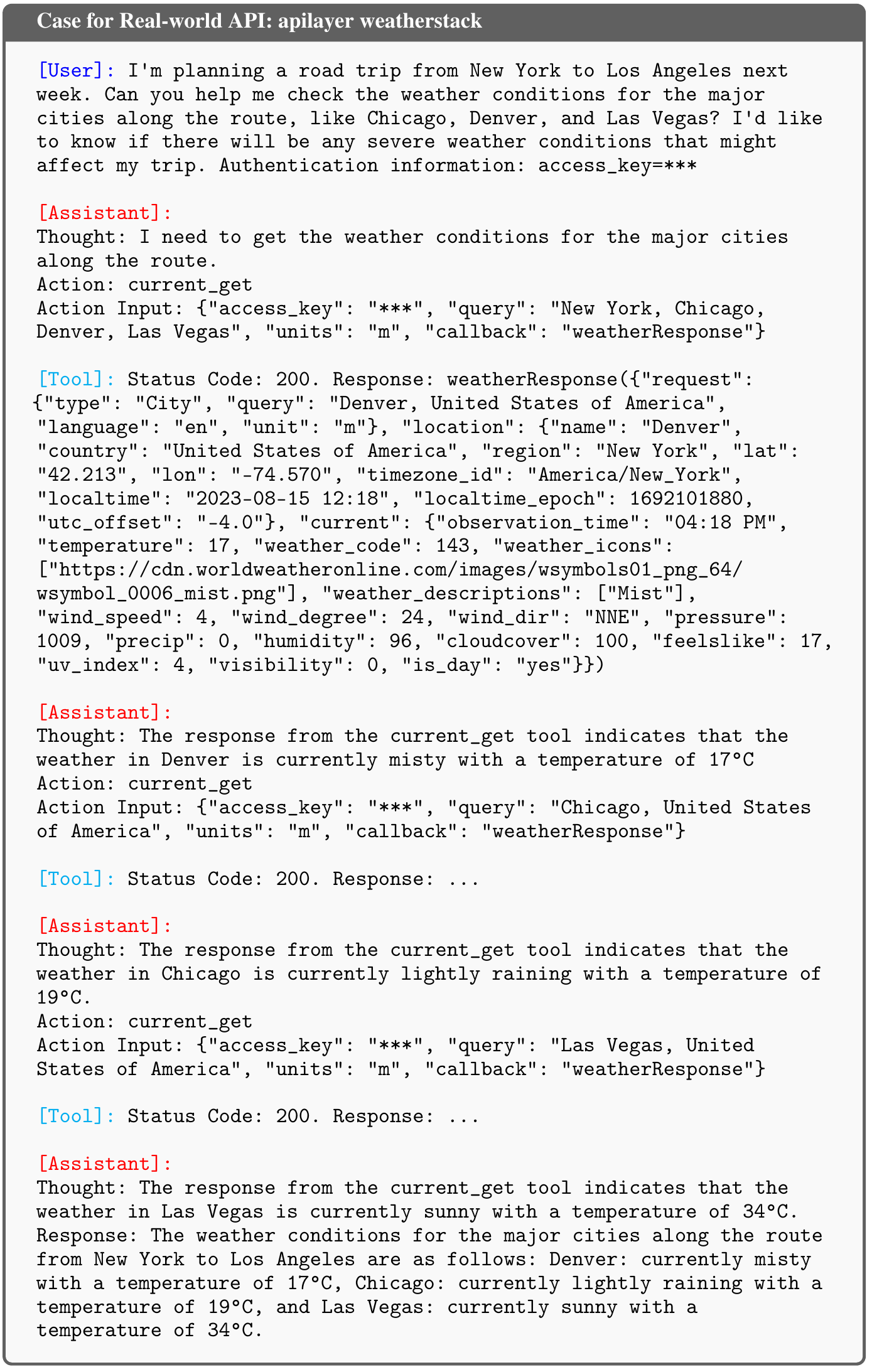}
  \caption{Cases for real-world API: apilayer weatherstack.}
  \label{fig:case3}
\end{figure*}

Through training on a set of diverse simulated tool-use instances, ToolAlpaca can equip various tools, even real-world APIs, some selected cases are shown in Figure~\ref{fig:case1}, Figure~\ref{fig:case2} and Figure~\ref{fig:case3}.

